\documentclass[letterpaper, 10 pt, conference]{ieeeconf}% Comment this line out if you need a4paper
\usepackage{amsmath,amsfonts,amssymb}
\usepackage{algorithmic}
\usepackage{array}
\usepackage{textcomp}
\usepackage{stfloats}
\usepackage{url}
\usepackage{xcolor}
\definecolor{gain}{RGB}{34,139,34} % 绿色
\definecolor{loss}{RGB}{255,0,0}% 橙红色
\usepackage{float}
\usepackage{verbatim}
\usepackage{breqn}
\usepackage{subcaption}
\usepackage{graphicx}
\usepackage{multirow} 
\usepackage{gensymb}
\usepackage{makecell}
\usepackage{siunitx} % 科学计数
\usepackage{booktabs}% 专业表格线
\usepackage{orcidlink}

\IEEEoverridecommandlockouts % Ths command is only needed if 
\title{\bfseries
HE-VPR: Height Estimation Enabled Aerial Visual Place Recognition Against Scale Variance
}

\author{
Mengfan He$^{1, \orcidlink{0009-0001-0053-2504}}$, 
Xingyu Shao$^{1, \orcidlink{0000-0002-2991-5883}}$, 
Chunyu Li$^{2, \orcidlink{0000-0002-1166-1555}}$, 
Chao Chen$^{1, \orcidlink{0009-0006-7391-9503}}$, 
Liangzheng Sun$^{3, \orcidlink{0009-0001-5899-8905}}$, \\
Ziyang Meng$^{1, \orcidlink{0000-0002-3742-0039}}$, 
and Yuanqing Wu$^{4, \orcidlink{0000-0001-9509-2670}}$% <- 换行符加在这里，将最后两位作者移至下一行
\thanks{*This work was supported in part by the Tsinghua-Toyota Joint Research Fund, in part by the Beijing Natural Science Foundation (Grant No.
L252095), and in part by the National Natural Science Foundation of China (Grant Nos. 62273195 and 62403269). \textit{(Mengfan He and Xingyu Shao are co-first authors.) (Corresponding author: Ziyang Meng.)}}%
\thanks{$^{1}$Mengfan He, Xingyu Shao, Chao Chen, and Ziyang Meng are with the Department of Precision Instrument, Tsinghua University, Beijing 100084, China (e-mail: hmf21@mails.tsinghua.edu.cn; shao-xy21@mails.tsinghua.edu.cn; chen-c@mail.tsinghua.edu.cn; ziyangmeng@mail.tsinghua.edu.cn).}%
\thanks{$^{2}$Chunyu Li is with the School of Aerospace Engineering, Beijing Institute of Technology, Beijing 100081, China (e-mail: chunyuli@bit.edu.cn).}%
\thanks{$^{3}$Liangzheng Sun is with the School of Instrumentation Science and Opto-electronics Engineering, Beijing Information Science and Technology University, Beijing 100192, China (e-mail: 2023030031@bistu.edu.cn).}%
\thanks{$^{4}$Yuanqing Wu is with the School of Intelligent Systems Engineering, Sun Yat-sen University, Guangzhou 510275, China (e-mail: wuyuanqing@mail.sysu.edu.cn).}%
}

\vspace{-2.0ex}

\begin{document}
\bstctlcite{MyBSTcontrol}

\maketitle
\thispagestyle{empty}
\pagestyle{empty}

\begin{abstract}
In this work, we propose HE-VPR, a visual place recognition (VPR) framework that incorporates height estimation.
Our system decouples height inference from place recognition, allowing both modules to share a frozen DINOv2 backbone.
Two lightweight bypass adapter branches are integrated into our system.
The first estimates the height partition of the query image via retrieval from a compact height database, and the second performs VPR within the corresponding height-specific sub-database.
The adaptation design reduces training cost and significantly decreases the search space of the database.
We also adopt a center-weighted masking strategy to further enhance the robustness against scale differences.
Experiments on two self-collected challenging multi-altitude datasets demonstrate that HE-VPR achieves up to 6.1\% Recall@1 improvement over state-of-the-art ViT-based baselines and reduces memory usage by up to 90\%.
These results indicate that HE-VPR offers a scalable and efficient solution for height-aware aerial VPR, enabling practical deployment in GNSS-denied environments.
All the code and datasets for this work have been released on \url{https://github.com/hmf21/HE-VPR}.
\end{abstract}

\section{Introduction}
\label{sec:intro}

Visual place recognition (VPR) in aerial platforms enables stable self-positioning in global navigation satellite system (GNSS)-denied environments. 
Such capabilities are critical for unmanned aerial vehicles (UAVs) facing severe accumulated dead-reckoning drift, necessitating absolute pose and scale re-initialization solely from vision.
However, varying flight altitudes cause significant scale variations that severely degrade the accuracy of place retrieval.
Since the visual footprint is strictly determined by the relative distance to the ground, we denote this scale-determining factor (i.e., relative altitude) as \textit{height} throughout this paper to ensure terminology consistency.
Since full-database retrieval covering all possible heights is infeasible for memory-constrained systems, estimating the flight height to explicitly narrow down the search space is essential.
While monocular metric depth estimation (MMDE) approaches~\cite{yangDepthAnythingV22024, piccinelliUniDepthV2UniversalMonocular2025, bhatZoeDepthZeroshotTransfer2023} can provide depth priors, such dense estimation solutions struggle on aerial platforms due to scarce annotations at large heights and weak local textures~\cite{shao2026altitudeawarevisualplacerecognition}.
To address these challenges, we propose \textit{HE-VPR}, an efficient framework leveraging a frozen DINOv2 backbone~\cite{oquabDINOv2LearningRobust2024} equipped with two parallel branches of lightweight bypass adapters. Specifically, each Transformer block is paired with two separate adapters dedicated to height estimation and VPR, respectively.

\begin{figure}[t]
\centering
\includegraphics[width=\linewidth]{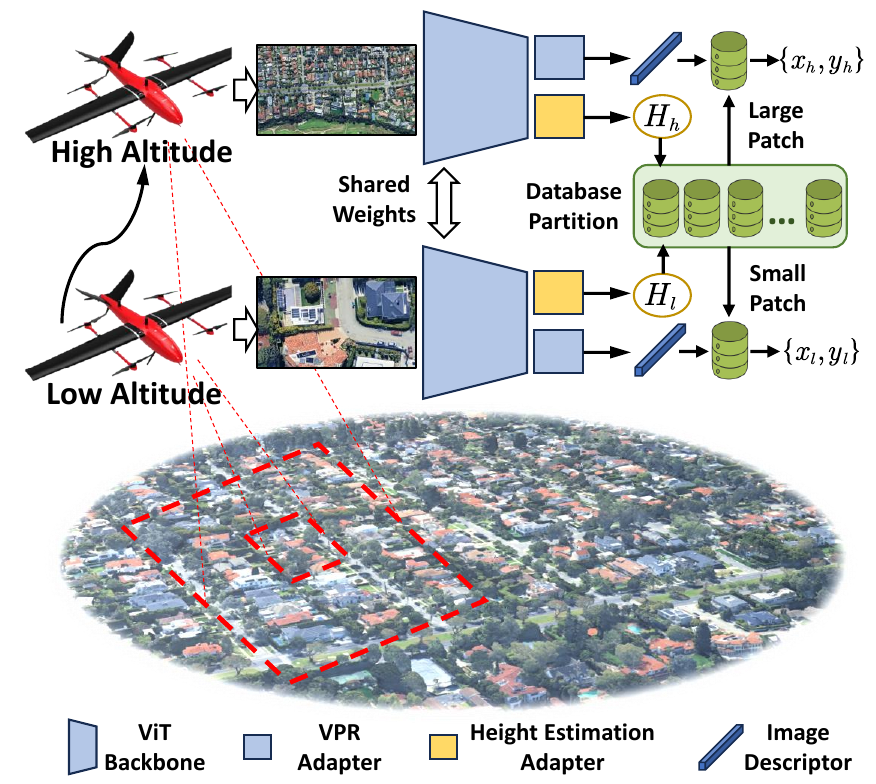}
\vspace{-3.0ex}
\caption{Overview of the proposed HE-VPR, an aerial VPR pipeline with height estimation.}
\vspace{-3.0ex}
\label{fig:paper_overview}
\end{figure}

In the first stage, the height adapter branch coarsely estimates the query image’s height via retrieval rather than direct regression, efficiently narrowing the search space to the correct sub-database for better generalization and lower memory consumption.
In the second stage, the VPR adapter branch performs retrieval within the selected sub-database. To handle the remaining scale variations within the discrete height partition, we employ a center-weighted masking strategy. Since peripheral features are highly susceptible to truncation during height changes, the central region inherently maintains higher visual overlap. Therefore, this strategy amplifies the geometrically stable central pixels while suppressing side features.
Furthermore, unlike existing in-block adapter designs (e.g., CricaVPR~\cite{luCricaVPRCrossimageCorrelationaware2024}), our side-branch adapters avoid interfering with the backbone’s features. This architecture restricts gradient flow exclusively within small side branches, significantly reducing training overhead. 
Together, these methods enable reliable and resource-efficient retrieval under dynamic height variations.
In summary, our main contributions are as follows:

\begin{enumerate}
\item We introduce a two-stage pipeline that coarsely estimates the UAV's height partition and performs retrieval in the corresponding sub-database, decoupling scale variance to reduce search costs without compromising overall accuracy.
\item We integrate two independent parallel branches of bypass adapters across the shared ViT blocks. This design prevents feature interference between height estimation and descriptor extraction while retaining the backbone's generalization capability with minimal parameters.
\item We propose a center-weighted feature masking strategy for the VPR branch. By prioritizing central features less prone to FOV truncation, it mitigates residual scale variations and improves retrieval reliability within the selected height partition.
\end{enumerate}

We validate the approach on two challenging multi-height UAV datasets. Results demonstrate that our \textit{HE-VPR} system drastically reduces memory usage (by up to 90\%) while maintaining comparable retrieval performance against full-database baselines. Ablations verify the benefits of the height estimation pipeline, advancing height-aware VPR toward practical deployment on aerial platforms.

\section{Related Works}
\label{sec:related_works}

\subsection{Aerial VPR with Height Variation}
\label{sec:review_VPR}

% Early works on aerial VPR demonstrated the feasibility of deep-learning-based aerial image matching~\cite{amerConvolutionalNeuralNetworkBased2017a, simonyan2015deepconvolutionalnetworkslargescale, patelVisualLocalizationGoogle2020}.
% Subsequent methods advanced UAV-to-satellite retrieval through sophisticated feature aggregation and semantic reasoning~\cite{mengAirGeoNetMapGuidedVisual2024, liuNovelEAGLeFramework2025, sunF3netMultiviewScene2023, guoFromSatelliteGround2024}, or by leveraging foundation models for universal VPR~\cite{keethaAnyLocUniversalVisual2023}.
% While some works also address dynamic environments~\cite{gurguVisionBasedGNSSFreeLocalization2022, lomoMultiMapVisual2025}, these methods largely assume a relatively stable scale relationship or fixed flight height.

Early aerial VPR demonstrated the feasibility of deep-learning-based matching~\cite{amerConvolutionalNeuralNetworkBased2017a, simonyan2015deepconvolutionalnetworkslargescale, patelVisualLocalizationGoogle2020}. Subsequent research advanced UAV-to-satellite retrieval via sophisticated feature aggregation~\cite{mengAirGeoNetMapGuidedVisual2024, liuNovelEAGLeFramework2025, sunF3netMultiviewScene2023, guoFromSatelliteGround2024} or foundation models~\cite{keethaAnyLocUniversalVisual2023}. Despite efforts in changing environments~\cite{gurguVisionBasedGNSSFreeLocalization2022, lomoMultiMapVisual2025}, most methods assume stable scale relationships or fixed flight heights.

% Unlike ground-based scenarios~\cite{toftLongTermVisualLocalization2022, lynenLargescaleRealtimeVisual2020}, aerial VPR faces severe height variations. 
% Handling scale differences via brute-force retrieval across a massive multi-scale database is computationally prohibitive for memory-constrained UAVs. 
% Alternatively, one can explicitly estimate height prior to retrieval.
% However, multi-view methods~\cite{khattakVisionBased3D2024} rely on continuous frames, making them inapplicable to single-image VPR queries. 
% Similarly, universal monocular metric depth estimation (MMDE) models~\cite{yangDepthAnythingV22024, piccinelliUniDepthV2UniversalMonocular2025, bhatZoeDepthZeroshotTransfer2023} predict dense depth but fail drastically at large UAV heights due to limited operational ranges and weak near-nadir textures~\cite{shao2026altitudeawarevisualplacerecognition}.

Unlike ground-based localization~\cite{toftLongTermVisualLocalization2022, lynenLargescaleRealtimeVisual2020}, aerial VPR must manage severe height variations. Brute-force multi-scale retrieval is computationally prohibitive for UAVs. While explicit height estimation can narrow search spaces, multi-view approaches~\cite{khattakVisionBased3D2024} require continuous sequences, and monocular metric depth estimation (MMDE) models~\cite{yangDepthAnythingV22024, piccinelliUniDepthV2UniversalMonocular2025, bhatZoeDepthZeroshotTransfer2023} often fail at significant heights due to weak nadir textures~\cite{shao2026altitudeawarevisualplacerecognition}.

% To address this, Shao et al.~\cite{shao2026altitudeawarevisualplacerecognition} proposed a two-stage pipeline separating height classification and place retrieval. 
% While effective, deploying disjoint heavy networks for these tasks incurs significant computational overhead.
% Building upon this paradigm, we decouple height estimation and descriptor extraction into two independent parallel bypass adapter branches sharing a single frozen backbone. 
% By reformulating height estimation as a lightweight retrieval task, this design prevents task interference and seamlessly selects the height-specific sub-database, retaining generalization without full fine-tuning.

Shao et al.~\cite{shao2026altitudeawarevisualplacerecognition} addressed these variations via a two-stage classification and retrieval pipeline. However, their reliance on disjoint heavy networks incurs significant computational overhead. Building on this paradigm, we reformulate height estimation as a lightweight retrieval task within a shared-backbone architecture. This design maintains task independence and enables efficient sub-database selection while drastically reducing the resource footprint compared to previous multi-model solutions.

\subsection{Adapters for Vision Transformers}
\label{sec:review_adapter}

Leveraging pre-trained foundation models (e.g., DINOv2~\cite{oquabDINOv2LearningRobust2024}) benefits aerial VPR, but full fine-tuning is computationally prohibitive. 
Adapters provide a parameter-efficient alternative by updating only small bottleneck modules while freezing the backbone~\cite{houlsbyParameterEfficientTransferLearning2019}, significantly reducing trainable parameters and memory footprint. 
Recently, CricaVPR~\cite{luCricaVPRCrossimageCorrelationaware2024} and SelaVPR~\cite{luSeamlessAdaptationPretrained2024} introduced adapters to the VPR domain to enhance cross-image correlation and semantic localization.

Architecturally, adapters are categorized by placement. 
\textit{(i) In-block adapters} are sequentially inserted within Transformer blocks, typically after multi-head attention (MHA) or feed-forward neural network (FFN) sublayers, as adopted by CricaVPR and SelaVPR. 
While yielding strong single-task adaptation, they force the main data stream to carry task-specific modifications, making them suboptimal for multi-task scenarios.
\textit{(ii) Bypass (side-branch) adapters} operate in parallel to the main backbone, merging outputs via residual connections. 
Crucially for \textit{HE-VPR}, bypass adapters allow multiple task-specific branches (height estimation and VPR extraction) to share a frozen backbone without feature interference. 
This high modularity enables our two-stage height-aware retrieval pipeline without the overhead of deploying separate heavy networks.

\section{Methodology}

\begin{figure}[b]
\centering
\includegraphics[width=\linewidth]{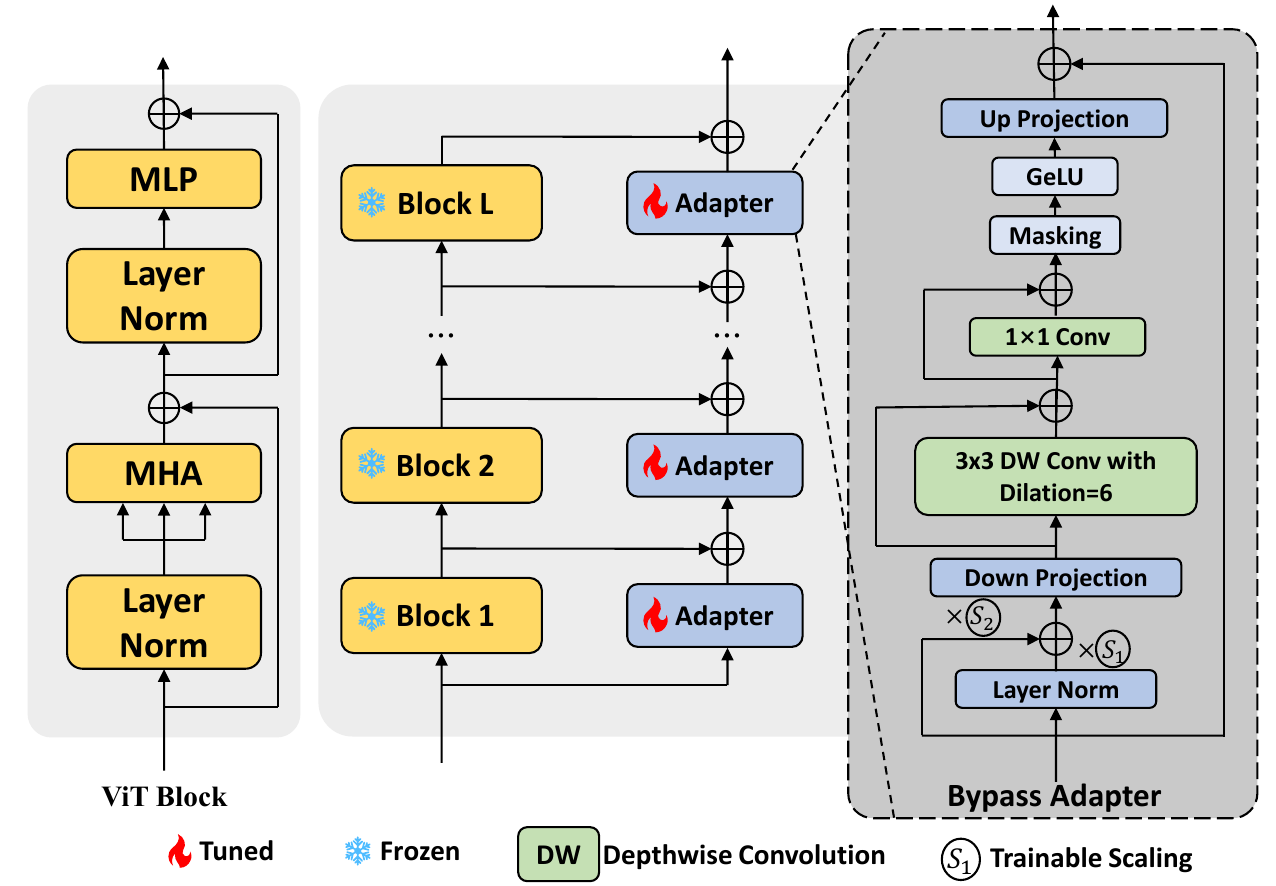}
\vspace{-2.0ex}
\caption{
Illustration of the adapter network in ViT.
}
\vspace{-1.0ex}
\label{fig:adapter_overview}
\end{figure}

\begin{figure*}[t]
\centering
\includegraphics[width=0.9\linewidth]{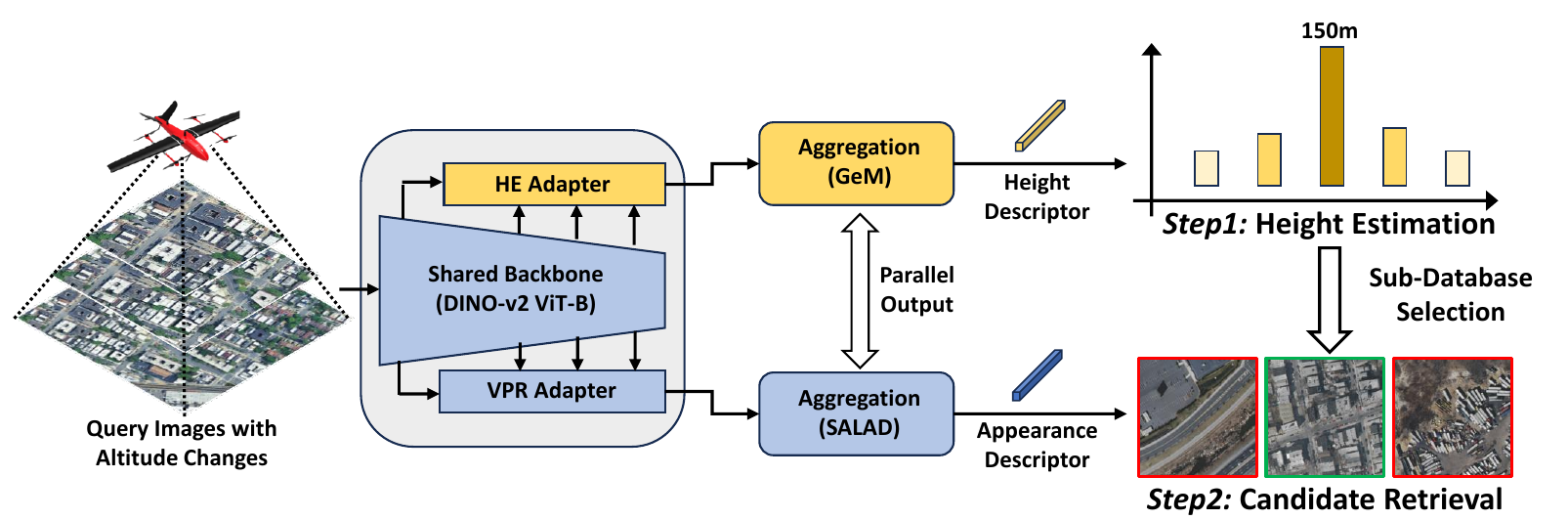}
\vspace{-1.0ex}
\caption{
\textbf{The proposed HE-VPR pipeline.}
A height estimation branch is added for sub-database selection such that the proposed pipeline is robust to height variance. It requires only a single forward pass of the model for both parts.
}
\vspace{-1.0ex}
\label{fig:VPR_artecheture_overView}
\end{figure*}

In this paper, we focus on the aerial VPR problem under significant height variance.
To address this challenge, we propose \textit{HE-VPR}, a two-stage retrieval pipeline that explicitly decouples height inference from place recognition.
As illustrated in Fig.~\ref{fig:VPR_artecheture_overView}, our framework leverages a shared frozen foundation model equipped with two independent, parallel adapter branches: the height estimation branch and the VPR branch.
For clarity and alignment with the illustrations in Fig.~\ref{fig:VPR_artecheture_overView}, the entire sequence of bypass adapters comprising these two paths is denoted as the \textit{HE adapter branch} and the \textit{VPR adapter branch}, respectively.
The \textit{HE adapter branch} extracts a coarse height descriptor to select the appropriate height-specific sub-database, while the \textit{VPR adapter branch} extracts the appearance descriptor for the final place retrieval.
The details of these components are formulated in the subsequent sections.

\subsection{Preliminary}
First, we provide a brief overview of the vision Transformer (ViT) and its adaptation mechanism for fine-tuning.
The ViT slices a given image into $N$ patches and projects them into $D$-dimensional embeddings, denoted as $x_p \in \mathbb{R}^{N \times D}$.
A learnable class token $x_{\text{class}} \in \mathbb{R}^{1 \times D}$ is concatenated with these embeddings, and the resulting sequence $x_0 = [x_{\text{class}}, x_p] \in \mathbb{R}^{(N+1) \times D}$ is fed into $L$ cascaded Transformer blocks to generate a discriminative feature representation $x_l$ (denoted as the output of the $l$-th Transformer block, where $l \in \{1, \dots, L\}$).
As shown in Fig.~\ref{fig:adapter_overview}, each Transformer block consists of a MHA module, a multilayer perceptron (MLP), and layer normalization (LN).

To perform task-specific fine-tuning without updating the backbone, bypass adapter modules are placed in parallel with the Transformer blocks.
Crucially, in our dual-branch architecture, each block $l$ is paired with two separate adapters to process the height estimation and place recognition tasks independently.
Let $t \in \{\text{he}, \text{vpr}\}$ denote the task index. The data flow within the adapter branch $t$ at the $l$-th block is formulated as:
\begin{equation}
\small
x_{l, t}^{(a)} = \text{Adapter}_t(x_{l-1} + x_{l-1, t}^{(a)}), \quad x_{0, t}^{(a)} = 0,
\end{equation}
where $x_{l, t}^{(a)}$ and $x_{l-1, t}^{(a)}$ are the outputs of the task-specific adapter at the $l$-th and $(l-1)$-th blocks, respectively.
In addition, $x_{l-1}$ represents the frozen intermediate feature extracted from the shared main backbone.

\subsection{Bypass Adapter}
Following the design of Mona~\cite{yin5>100BreakingPerformance2025}, we embed a bypass adapter for both the HE and VPR branches within each Transformer block. To ensure the computational efficiency of these bypass adapters, we employ only a single depth-wise convolution with a 3$\times$3 kernel.
We use dilated convolution to increase the kernel size, which ensures the adapter network has a larger receptive field while maintaining efficiency.
Such a design also aligns with the large-scale feature information present in aerial imagery.
Corresponding to the right subplot of Fig.~\ref{fig:adapter_overview}, the overall calculation within the adapter can be formulated as follows:
\begin{dmath}
\small
x_{l,t}^{(a)} = x_{0} + U \sigma \left( \mathcal{M} \left( f_{pw} \left( f_{dw} \left( D \left( s_1 \|x_0\| + s_2 x_0 \right) \right) \right) \right) \right),
\end{dmath}
where $x_0 = x_{l-1} + x_{l-1, t}^{(a)}$ is the input to the adapter, derived from the summation of the shared backbone feature $x_{l-1}$ and the task-specific adapter feature $x_{l-1, t}^{(a)}$.
$D$ and $U$ are the down-projection and up-projection linear layers, while $\sigma$ denotes GeLU activation.
In addition, $f_{pw}$ and $f_{dw}$ are the point-wise and depth-wise convolution operations containing shortcuts, respectively, and $\mathcal{M}$ is the masking strategy applied to the extracted features.

These bypass adapters are designed to operate exclusively within a parallel path, receiving outputs from preceding blocks without modifying the subsequent data flow in the main backbone. This architecture is a simple yet essential design, as it allows us to freeze the parameters of the pre-trained foundation model and exclusively train the adaptation networks. Furthermore, it enables the parallel branches to operate independently with isolated data streams. Leveraging this design, we construct two independent network branches dedicated to the HE and VPR tasks, respectively.

\subsection{Height-Aware Aerial VPR}
\label{sec:method_HE_adapter}
Traditional VPR methods designed for multi-height applications can only handle limited scale variations (as mentioned in Sec.~\ref{sec:review_VPR}).
When facing height variance in a large range, such as during takeoff and landing, existing VPR methods fail to maintain robustness.
To address this, we propose \textit{HE-VPR}, a novel paradigm for height-aware aerial VPR, as shown in Fig.~\ref{fig:VPR_artecheture_overView}.
Following the standard VPR offline preparation process, we first construct a multi-level map database to accommodate flights at varying heights.
Each height level corresponds to a specific sub-database. To establish a geometric relationship between the flight height and the map scale, the physical height ranges are mapped to the image's spatial footprint using camera intrinsics. This allows for a logical partitioning of the database:

\begin{dmath}
\small
\mathbf{D} = \underbrace{\{\mathbf{D}^1,\cdots, \overbrace{\mathbf{D}^l=\{\mathbf{d}_1^l, \mathbf{d}_1^l, \dots, \mathbf{d}_n^l}^{n\text{ samples in } l\text{-th sub-database }}\}, \dots, \mathbf{D}^L\}}_{L\text{ levels of subsets in the database}},
\end{dmath}
where $\mathbf{D}^l$ and $\mathbf{d}_i^l$ represent the sub-databases and individual map samples, respectively.
To ensure all flight heights are covered, the $L$ levels of sub-databases are established based on the range of heights encountered.
In a one-stage VPR pipeline, direct retrieval from the aggregated database $\mathbf{D}$ would result in significant computational and memory overhead, making edge-device implementation infeasible.
Leveraging the previously independent adapter designs, we decompose this task into a two-step process: sub-database selection and retrieval, handled by two independent adapter branches, respectively.

For the height estimation branch, precise height values are not required, as height is only utilized for discrete sub-database selection.
Therefore, we reformulate height estimation as a retrieval task, similar to VPR.
We use a simple GeM~\cite{radenovicFinetuningCNNImage2018} pooling layer to extract a height descriptor.
A compact database for height retrieval is collected from map patches at different scales.
The height information is obtained through a retrieval-based approach rather than direct regression.
This enables more convenient generalization to new environments by changing the corresponding height database.
Furthermore, since the database is expandable, we can leverage top-$k$ candidates (e.g., top-5) to select multiple sub-databases, enhancing selection accuracy compared to direct depth estimation pipelines.
The minimal overhead introduced by the compact height database allows us to maintain high retrieval efficiency, especially when compared to the case of a densely sampled full database.

In the VPR branch, we follow the classical feature extraction and aggregation pipeline to ensure retrieval performance.
We use a modified Mona network~\cite{yin5>100BreakingPerformance2025} as the adapter and utilize SALAD~\cite{izquierdo2024optimal} as the aggregation module.
The extracted image descriptors are retrieved against the corresponding database, which is selected by the height estimation branch.
Because only the simplest GeM layer is used for the height descriptor and the foundation model is shared by both branches, this approach only causes a slight increase in computational costs compared to a classical VPR approach.
Two kinds of descriptors are extracted simultaneously and independently of each other, meaning they can be trained individually.
Therefore, we can train these two branches with different datasets, the details of which will be given in Sec.~\ref{sec:implementation}.

\subsection{Masked Feature Enhancement}
Considering the height estimation stage, although we can obtain the correct sub-database selection, the discrete partitioning still causes a certain degree of scale difference between the query image and the referenced database image.
As mentioned before, the design of one adapter branch is isolated and does not impact the other.
Therefore, we can further design a masking strategy within the VPR adapter branch to provide better adaptability to these residual small-scale height changes, as shown in Fig.~\ref{fig:adapter_overview}.
Combined with the height estimation adapter branch, this design allows for accurate visual place recognition from coarse to fine, even under significant height variations.

At the VPR bypass, we add a masking mechanism based on the feature variance after the point-wise convolution, which can be represented by the following formulation:

\begin{dmath}
\label{eq:mask}
\small
\mathcal{M}_{i,j} = \exp{\left(-\frac{(j - \frac{W}{2})^2 + (i - \frac{H}{2})^2}{2 \cdot (\frac{\max(H, W)}{2})^2}\cdot \text{Var}(x)\right)},
\end{dmath}
where $\mathcal{M}_{i,j}$ denotes the feature mask value at feature position $(i,j)$, and $H$, $W$ are the height and width of the feature map.
We calculate the feature variance of each channel, denoted as $\text{Var}(x)$.
Subsequently, the features are multiplied element-wise with the mask $\mathcal{M}$ before being passed through the non-linear activation function $\sigma$.

As presented in Eq.~\ref{eq:mask}, the feature mask is primarily determined by two factors: pixel position and the variance of the feature map.
This design explicitly addresses the geometric realities of downward-facing cameras during height changes. Under significant height variations, the central region of the downward-facing camera's field of view remains relatively content-stable, merely undergoing scale changes. Conversely, the edge regions are highly susceptible to disappearing (truncation as height decreases) or reappearing.
This formulation gives pixels closer to the image center higher activation values, mitigating the negative impact of unstable peripheral features.
Furthermore, we assume that a higher feature variance indicates a higher flight height, where scale distortion effects are more pronounced.
For such feature maps, we reduce the contribution of their edge pixels to the final result, focusing instead on the more stable central region.
By using feature masking, we force the VPR adapter to focus on the central region of the feature map, making the retrieval process more robust to the remaining scale changes within the selected height partition.

\section{Experiments}
\label{sec:evaluation}

\subsection{Implementation Details}
\label{sec:implementation}
We use ViT-Base DINOv2 pre-trained model~\cite{oquabDINOv2LearningRobust2024} as the foundation backbone and the two side branches are formulated as adapter branches as presented in Fig.~\ref{fig:adapter_overview}.
The resolution of the input image is $224 \times 224$ in both training and inference stages.
The token dimension in the backbone is 768 and the feature dimension in the bypass adapters is 64.
The VPR adapter branch and other evaluated methods are all retrained on the same dataset proposed by He et al.~\cite{he2024leveraging}, which consists of satellite maps from multiple years. % TODO:这里应该更新成新的数据集
The HE adapter branch is trained on the same dataset as the VPR adapter branch by resizing the satellite images to various scales.
As the two adapter branches both follow the metric learning pipeline, they share almost identical architectural configurations and parameter sizes.
We train our models using the AdamW optimizer with the initial learning rate set as 0.00001 and the multi-similarity loss~\cite{ali2022gsv}.
A training batch contains 32 classes with 2 images each.
The models are trained for a sufficient number of epochs, and the weights that achieve the highest performance on the validation set are selected.
All the training experiments are deployed on an NVIDIA 4090 GPU with the same framework proposed in MixVPR~\cite{beyMixVPRFeatureMixing2023} using PyTorch.

We evaluate the VPR performance with the commonly used Recall$@$N (R$@$N) metric, which is the percentage of queries that have the correct result among the N retrieved images.
As the evaluation datasets contain various image scales, the distance threshold for correct place retrieval also varies.
Therefore, we present the performance at different positive thresholds to evaluate the correct retrieval across various scales.
Given that the HE adapter branch follows the retrieval pipeline, we also provide the evaluation with R$@$N.
In order to evaluate the algorithm efficiency, we report the memory usage and the number of parameters.
All the evaluations are conducted on an NVIDIA 4060 GPU.

\subsection{Evaluation Datasets}
\label{sec:evaluation_datasets}
\begin{figure}[t]
\centering
\begin{subfigure}{\linewidth}
\centering
\includegraphics[width=0.8\linewidth]{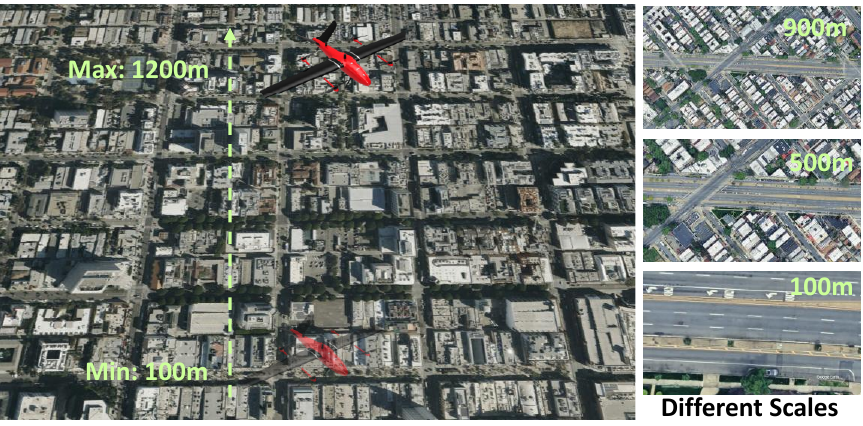}
\vspace{-1.0ex}
\caption{GEStudio Dataset: urban area with buildings and roads.}
\label{fig:GEStudio_sample}
\end{subfigure}
\begin{subfigure}{\linewidth}
\centering
\includegraphics[width=0.8\linewidth]{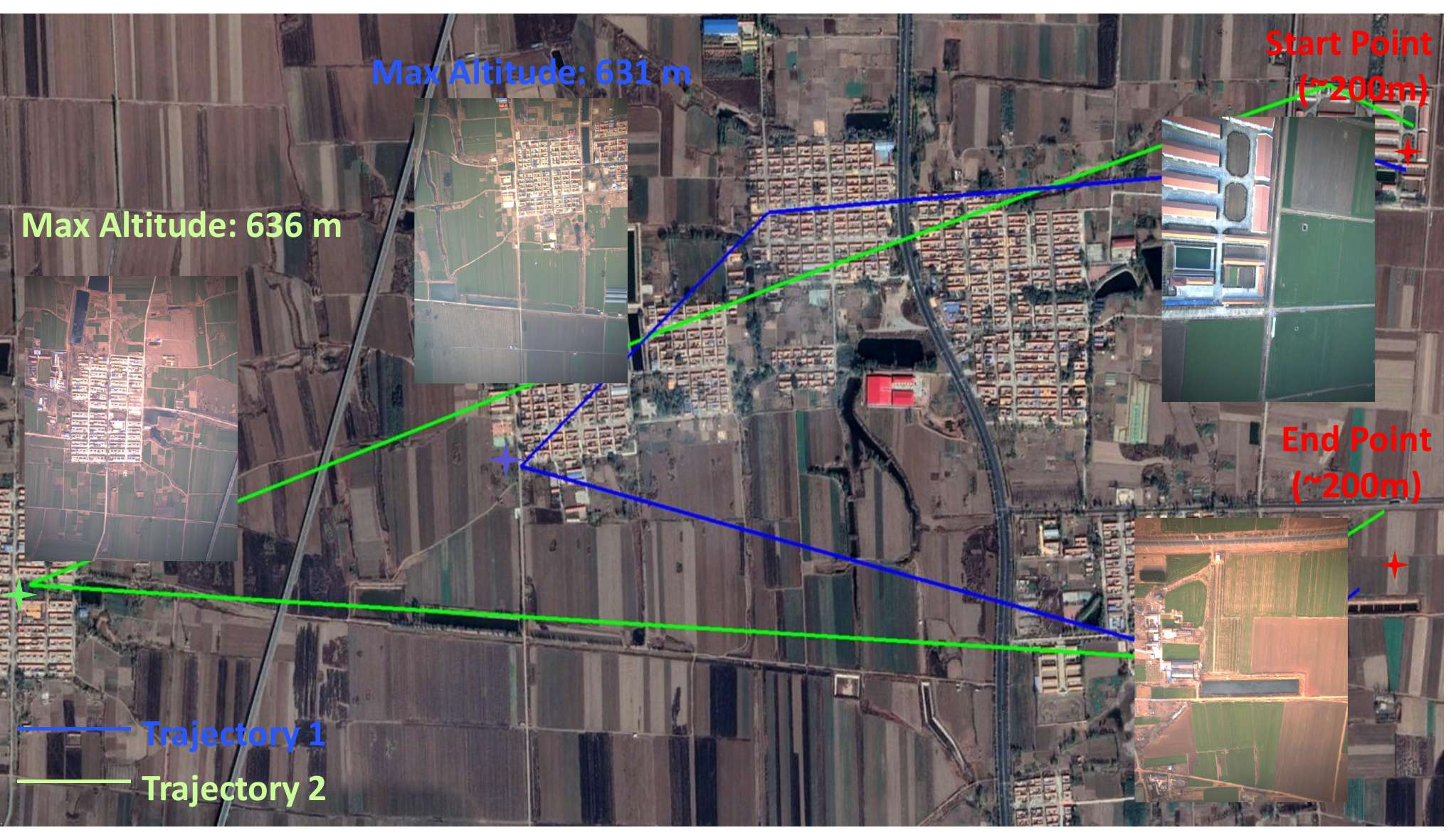}
\vspace{-1.0ex}
\caption{MHFlight Dataset: rural area with farmland and ponds.}
\label{fig:MHFlight_sample}
\end{subfigure}
\vspace{-2.0ex} % 调整子图间距
\caption{Overviews of two evaluation datasets.}
\vspace{-2.0ex}
\label{fig:dataset_overview}
\end{figure}

Since the proposed height estimation module is capable of accommodating height variance of several kilometers, the evaluation datasets should ideally contain a sufficiently wide range of height changes.
Several commonly-used aerial VPR datasets such as University-1652~\cite{zheng2020university}, SUES-200~\cite{zhu2023sues}, DenseUAV~\cite{dai2023vision} only contain data with a height difference no more than two hundred meters and are therefore not enough for our experiments.
Therefore, we provide two self-collected datasets with larger height variance that are more suitable for the evaluation of HE-VPR and the details are shown as follows:

\textit{1) GEStudio} is a simulated dataset of drone imagery collected from Google Earth Studio, which can provide animated imagery content.
We sample 1200 images from 100 locations in New York, U.S., with heights ranging from 100 to 1200 \si{\m}.
The images in this dataset feature significant variations in scale from near-ground to height over 1000 \si{\m}, and Fig. \ref{fig:GEStudio_sample} illustrates the sampling process at one of the selected locations.
\textit{2) MHFlight} (Multi-Height Flight) is a real-world dataset collected from drone flights. We collect this dataset in a rural village, with varying flight heights between 200 \si{\m} and 600 \si{\m}. % Jimo, China
In contrast to GEStudio dataset, which primarily features urban scenes, MHFlight dataset mostly covers agricultural areas.
Another difference is that the heights of MHFlight dataset vary continuously, whereas the height in the previous dataset changes discretely.
MHFlight dataset comprises 1970 images captured from two distinct flight trajectories, as shown in Fig. \ref{fig:MHFlight_sample}.

\begin{table}[t]
\centering
\footnotesize
\caption{The statistics of GEStudio and MHFlight.}
\label{tab:datasets_info}
\renewcommand{\arraystretch}{1.3}
\begin{tabular}{l|cccc}
% \toprule
\noalign{\hrule height 1pt}
\multirow{2}{*}{Dataset} & \multirow{2}{*}{\parbox{1cm}{\centering Query Image}} & \multirow{2}{*}{\parbox{1cm}{\centering Height Database}} & \multirow{2}{*}{\parbox{1cm}{\centering VPR Database}} & \multirow{2}{*}{\parbox{1.3cm}{\centering Height Range (\si{\m})}} \\
 &&&&\\
\hline
\hline
GEStudio & 1200 & 102 & 29768 & 100$\sim$1200 \\
\hline
MHFlight & 1970 & 850 & 36400 & 200$\sim$640 \\
% \bottomrule
\noalign{\hrule height 1pt}
\end{tabular}
\vspace{-2.0ex}
\end{table}

These two datasets are both utilized in the evaluations of the HE and VPR adapter branches, with detailed statistics provided in Tab.~\ref{tab:datasets_info}.
Notably, as both of these datasets are unseen during training, they provide a direct validation of the model ability for zero-shot generalization.
All of these query images are paired with their corresponding satellite maps, which are also collected from the Google Earth.
Following the multi-level map database strategy mentioned before, we create a map database composed of map tiles of various sizes, with each group representing a 50-meter interval.
The height information in the two evaluation datasets is converted into image widths using the intrinsic camera parameters, and then the images can match with a sub-database according to the width.

\subsection{HE Evaluation}

\begin{table}[t]
\centering
\footnotesize
\caption{Evaluation for the height estimation. We present the performance with height thresholds of 50m and 100m to define a correct retrieval. We compare with two MMDE methods, namely UniDepth V2 and Depth Anything (DA) v2. The best result is highlighted in bold.}
\label{tab:height_estimation_comparison}
\renewcommand{\arraystretch}{1.4}
\begin{tabular}{l|ccc|c}
\noalign{\hrule height 1pt}
\multirow{2}{*}{Method} & R$@$1 & R$@$5 & R$@$10 & {E$_{\text{avg}}$}\\
 & \multicolumn{3}{c|}{threshold at 50 \si{\m} and 100 \si{\m} ($\%$/$\%$)} & (\si{\m}) \\
\hline
\hline
% \noalign{\hrule height 1pt}
% \noalign{\hrule height 1pt}
\multicolumn{5}{c}{GEStudio dataset} \\
\hline
UniDepth V2 & 1.25/8.50 & N/A & N/A & 470.83 \\
\hline
DA V2 & 5.58/15.00 & N/A & N/A & 550.03\\
\hline
{HE-VPR} & \textbf{63.08/91.75} & \textbf{92.33/99.92} & \textbf{98.25/100.00} & \textbf{43.63}\\
\hline
\hline
% \noalign{\hrule height 1pt}
% \noalign{\hrule height 1pt}
\multicolumn{5}{c}{MHFlight dataset} \\
\hline
UniDepth V2 & 0.00/1.37 & N/A & N/A & 213.24 \\
\hline
DA V2 & 0.00/3.86 & N/A & N/A & 310.55 \\
\hline
{HE-VPR} & \textbf{34.01/60.96} & \textbf{76.95/91.83} & \textbf{89.85/96.80} & \textbf{93.50} \\
\noalign{\hrule height 1pt}
\end{tabular}
\vspace{-3.0ex}
\end{table}

In this section, we report the performance of the proposed height estimation adapter branch on the aforementioned two datasets and the comparison with two state-of-the-art MMDE methods.
The range for a correct retrieval in height estimation is set to 50 meters and 100 meters.
We present the results using standard retrieval metrics: R$@$1, R$@$5 and R$@$10.
However, since MMDE methods directly estimate a single height value, they do not produce multiple ranked candidates.
Therefore, we report only the R$@$1 metric for these methods and use N/A for the R$@$5 and R$@$10 metrics in the results.
And we also compare the average estimation error (denoted as E$_{\text{avg}}$) for the evaluated methods.
The evaluation results are shown in Tab.~\ref{tab:height_estimation_comparison}.

Since traditional MMDE methods fail to consider the downwards perspective of UAV imagery, their depth estimation performance is almost entirely unacceptable on the two evaluation datasets.
Our method, which obtains image height information in a retrieval manner, demonstrates stronger performance.
The height estimation performs significantly well on GEStudio dataset, with the majority of images being correctly categorized.
In contrast, the performance on the MHFlight dataset is relatively lower, where the farmland imagery is prevalent.
However, given that the flight heights in this dataset are all above 200 \si{\m}, a top-1 classification accuracy of 60.96\% with a 100 \si{\m} threshold is acceptable.
Regarding the average height estimation error E$_{\text{avg}}$, we use the label of the top-1 candidate as the estimation result, and the errors are 43.63 \si{\m} and 93.50 \si{\m} respectively.
Given that the height database is partitioned with a 50-\si{\m} interval, this level of model performance is considered acceptable.
Additionally, the role of height estimation differs from that of place recognition since top-5 or even top-10 candidates can be used to select sub-database as described in Sec.~\ref{sec:method_HE_adapter}.
Therefore, such height estimation performance proves to be entirely sufficient for VPR in the following stage.

\begin{figure}[t]
\centering
\includegraphics[width=\linewidth]{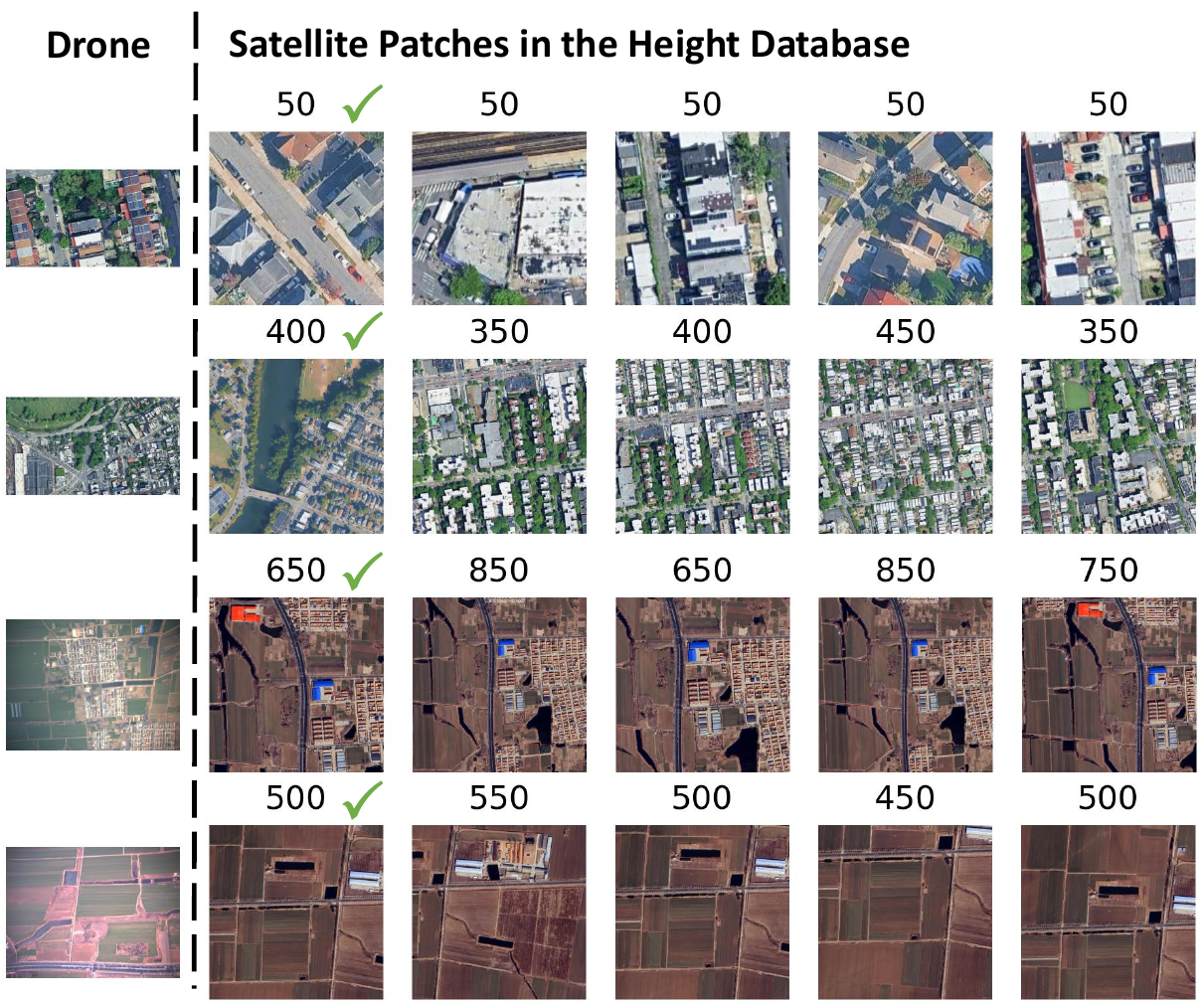}
\vspace{-1.0ex}
\caption{
Qualitative results for height estimation.
The top two rows are from GEStudio dataset and the bottom two rows are from MHFlight dataset.
The height label for top-5 database image is annotated at the top and all the samples are correctly retrieved at the top-1 candidate.
}
\vspace{-3.0ex}
\label{fig:HE_adapter_examples}
\end{figure}

We also present the top-5 database images retrieved by HE-VPR for four query examples from two evaluated datasets, as shown in Fig.~\ref{fig:HE_adapter_examples}.
As demonstrated by the examples, the proposed height estimation module is capable of recognizing the height changes of the captured samples and retrieve database images with similar heights.
It also indicates that the HE adapter branch can be used for samples at both very high and relatively low heights.
As shown in Fig.~\ref{fig:HE_adapter_examples}, the proposed HE adapter branch uses a height database that can be tailored to different datasets, which helps to ensure accurate height estimation.
This offers a key advantage over MMDE methods, as our module can leverage knowledge from the database, rather than relying on direct height regression.
By applying a suitable height database, the proposed HE adapter achieves strong generalization across a wide range of scenarios.
To validate this approach, we conduct an experiment on the effect of using different height databases, with the results presented in Tab.~\ref{tab:height_estimation_different_database}.
Although the HE adapter branch does not strictly require a certain database, there is a notable decrease using a different database compared to the performance achieved by the database with similar appearance.
Moreover, using a larger height database does not lead to better performance.
This suggests that, unlike the VPR task which benefits from densely sampled database, height estimation can be negatively impacted by a large number of database samples.
This is because the number of height labels is significantly smaller than in traditional VPR tasks, often no more than one hundred.
Consequently, establishing an appropriately sized database is critical for this task.

\begin{table}[b]
\centering
\footnotesize
\caption{Height estimation performance with different databases. In this evaluation, \textbf{G} and \textbf{M} denote denote databases sharing similar scene domains (e.g., urban and rural environments) with the GEStudio and MHFlight datasets, respectively. And the superscript $^*$ indicates a database that is larger than the original database.}
\label{tab:height_estimation_different_database}
\renewcommand{\arraystretch}{1.3}
\begin{tabular}{c|c|ccc}
% \toprule
\noalign{\hrule height 1pt}
{Dataset} & {Database} & R$@$1 & R$@$5 & R$@$10\\
\hline
\multirow{3}{*}{GEStudio} & \textbf{M} & 36.50/54.25 & 73.33/91.08 & 91.58/98.33 \\
\cline{2-5}
& \textbf{G} & \textbf{63.08/91.75} & \textbf{92.33/99.92} & \textbf{98.25/100.00}\\
\cline{2-5}
& \textbf{\hphantom{$^*$}G$^*$} & 61.58/91.33 & 90.25/99.75 & 95.58/99.92 \\
\hline
\hline
\multirow{3}{*}{MHFlight} & \textbf{G} & 29.04/49.75 & 52.08/67.26 & 66.29/72.13 \\
\cline{2-5}
& \textbf{M} & \textbf{34.01/60.96} & \textbf{76.95/91.83} & \textbf{89.85/96.80} \\
\cline{2-5}
& \textbf{\hphantom{$^*$}M$^*$} & 30.00/50.36 & 70.66/88.58 & 82.44/95.13 \\
\noalign{\hrule height 1pt}
\end{tabular}
\end{table}

\subsection{VPR Evaluation}
In this section, we compare the proposed scale-invariant VPR adapter branch on the proposed two datasets with a wide range of SOTA VPR methods.
We select methods from both CNN-based architectures (GeM~\cite{radenovicFinetuningCNNImage2018}, Cosplace~\cite{berton2022rethinking}, MixVPR~\cite{beyMixVPRFeatureMixing2023}) and ViT-based architectures (CricaVPR~\cite{luCricaVPRCrossimageCorrelationaware2024}, SALAD~\cite{izquierdo2024optimal}).
To guarantee a fair comparison, all the methods are retrained on the same dataset and use appropriate inference parameters.
In this part of the evaluation, we do not include the results of the HE adapter branch and only evaluate the VPR adapter branch based on the full database.
Therefore, the retrieval database contains all the sub-databases for different height levels.
As the two branches are isolated, we can independently evaluate the performance of the proposed VPR model.
In this part of the experiment, we select 100 \si{\m} and 200 \si{\m} as the positive thresholds.
In addition, we evaluate the number of parameters for each method, and the results are presented in Tab.~\ref{tab:VPR_adapter_evaluation}.

In summary, the proposed VPR adapter branch achieves outstanding performance, especially on the GEStudio dataset.
MHFlight dataset presents a significant challenge as it is largely comprised of low-texture areas.
Therefore, most methods fail to demonstrate strong performance. Nevertheless, our approach is still notable, achieving the best performance in the overall metrics.
Also, the VPR adapter branch outperforms the original DINOv2-based SALAD model across all the evaluated datasets.
This demonstrates that the adapter structure effectively improves the performance of VPR.
In terms of the number of parameters, our method adds 2.6 \si{MB} of parameters by introducing inter-block adapters.
The higher performance of MixVPR on the MHFlight dataset is partly due to its use of a higher image resolution (320$\times$320), which is larger than the 224$\times$224 resolution typically used by DINO-based methods.
Finally, the proposed modular design also allows for seamless integration of future advanced methods, enabling us to update individual components based on different applications without affecting the adapter branches within the system.

\begin{table*}[t]
\centering
\footnotesize
\caption{VPR comparison of different methods. We only use the VPR adapter branch for evaluation. Here we also present the performance under different thresholds with 100 \si{\m} and 200 \si{\m}. The best results are shown in bold.}
\label{tab:VPR_adapter_evaluation}
\renewcommand{\arraystretch}{1.3}
\begin{tabular}{l|c|c|c||ccc||ccc}
% \toprule
\noalign{\hrule height 1pt}
\multirow{2}{*}{Method} & \multirow{2}{*}{Backbone} & \multirow{2}{*}{Dim} & Param. & \multicolumn{3}{c||}{GEStudio dataset} & \multicolumn{3}{c}{MHFlight dataset} \\
\cline{5-10}
& & & (\si{MB}) & R$@$1 & R$@$5 & R$@$10 & R$@$1 & R$@$5 & R$@$10 \\
\hline
GeM~\cite{radenovicFinetuningCNNImage2018} & ResNet50 & 1024 & 8.5 & 20.08/24.00 & 35.75/42.33 & 43.08/51.67 & 37.31/63.81 & 61.88/75.69 & 69.34/78.88 \\
\hline
CosPlace~\cite{berton2022rethinking} & ResNet50 & 2048 & 27.7 & 47.92/51.50 & 61.42/66.50 & 66.17/71.83 & 55.89/83.35 & 67.61/84.62 & 72.99/84.87 \\
\hline
MixVPR~\cite{beyMixVPRFeatureMixing2023} & ResNet50 & 4096 & 10.9 & 50.75/53.92 & 61.08/65.67 & 66.00/70.92 & \textbf{58.63}/84.47 & 71.83/84.82 & 77.26/85.03 \\
\hline
CricaVPR~\cite{luCricaVPRCrossimageCorrelationaware2024} & DINOv2-B & 10752 & 106.8 & 61.17/65.58 & 73.25/76.75 & 76.42/79.67 & 57.21/83.20 & 70.00/84.37 & 76.19/84.47 \\
\hline
SALAD~\cite{izquierdo2024optimal} & DINOv2-B & 8448 & 88.0 & 63.42/67.00 & 73.17/76.67 & 76.33/79.75 & 53.15/83.45 & 67.16/85.13 & 72.49/85.23 \\
\hline
\hline
{HE-VPR} & DINOv2-B & 8448 & 90.6 & \textbf{69.50/71.25} & \textbf{76.42/78.17} & \textbf{79.08/81.50} & 57.61\textbf{/85.48} & \textbf{72.49/86.04} & \textbf{77.82/86.35}\\
\noalign{\hrule height 1pt}
\end{tabular}
\vspace{-2.0ex}
\end{table*}

\subsection{HE-VPR Evaluation}

We finally evaluate the overall HE-VPR system that integrates the height estimation adapter branch with the VPR adapter branch, with results summarized in Tab.~\ref{tab:HE_VPR_evaluation_1}. By incorporating height estimation, retrieval is restricted to a relevant sub-database rather than the entire gallery. For comparison, we also evaluate the VPR adapter branch alone using full-database retrieval, denoted as \textit{Adapter(full)}. As described in Sec.~\ref{sec:method_HE_adapter}, we further exploit not only the top-1 height estimate but also the top-5 and top-10 candidates, denoted as \textit{HE-VPR(1)}, \textit{HE-VPR(5)}, and \textit{HE-VPR(10)}, respectively. All results are reported at a 100~\si{\m} positive threshold for clarity.
Without height estimation, a multi-level sub-database covering different height ranges is required; otherwise, accuracy drops sharply, as shown in the first row of Tab.~\ref{tab:HE_VPR_evaluation_1}. Although a full-scale database contains all height information, it does not guarantee optimal performance in all scenarios. Height estimation enhances retrieval accuracy, but relying solely on the top-1 estimate can lead to performance loss due to estimation uncertainty. Selecting multiple top-$k$ candidates for sub-database retrieval mitigates this issue and is a key advantage of the proposed HE adapter branch over direct regression approaches.

\begin{table}[!t]
\centering
\footnotesize
\caption{Comparison between the standalone VPR adapter branch and the proposed HE-VPR system.}
\label{tab:HE_VPR_evaluation_1}
\renewcommand{\arraystretch}{1.4}
\begin{tabular}{l|ccc|ccc}
% \toprule
\noalign{\hrule height 1pt}
\multirow{2}{*}{Method} & \multicolumn{3}{c|}{GEStudio dataset} & \multicolumn{3}{c}{MHFlight dataset} \\
\cline{2-7}
& R$@$1 & R$@$5 & R$@$10 & R$@$1 & R$@$5 & R$@$10 \\
% \hline
% Adapter(part) & 32.50 & 50.50 & 58.75 & 34.87 & 51.73 & 57.06 \\
\hline
Adapter(full) & {69.50} & {76.42} & \textbf{79.08} & \textbf{57.61} & 72.49 & \textbf{77.82} \\
\hline
HE-VPR(1) & 57.25 & 66.50 & 70.00 & 49.14 & 68.07 & 74.06 \\
\hline
HE-VPR(5) & 69.92 & 76.17 & 78.67 & 56.80 & \textbf{72.94} & {77.72} \\
\hline
HE-VPR(10) & \textbf{70.42} & \textbf{76.83} & 79.00 & {57.41} & 71.93 & 77.46 \\
% \bottomrule
\noalign{\hrule height 1pt}
\end{tabular}
\end{table}

\begin{table}[!t]
\centering
\footnotesize
\caption{Comparison for memory usage and performance ratio of database.}
\label{tab:HE_VPR_evaluation_2}
% 1. 将全局行高倍数恢复到接近正常的水平 (1.1)
\renewcommand{\arraystretch}{1.1} 
\begin{tabular}{l|c|lll}
\noalign{\hrule height 1pt}
{Method} & {Full} & HE-VPR(1) & HE-VPR(5) & HE-VPR(10) \\
\hline
\hline
% 2. 数据集标题行：通过 \rule 控制一个小高度，保持紧凑
\multicolumn{5}{c}{\rule{0pt}{2.3ex} GEStudio dataset} \\ 
\hline
% 3. 数据行：使用 parbox[c][高度][c] 实现真正的垂直居中和行宽控制
\parbox[c][6ex][c]{1.8cm}{\raggedright Memory Usage ($\%$)} & 100 & \textbf{12.50} & 38.71 & 58.51 \\
\hline
\parbox[c][6ex][c]{1.8cm}{\raggedright Performance Ratio ($\%$)} & 100 & {86.11}$_{{\color{loss}-13.89}}$ & 99.89$_{{\color{loss}-0.11}}$ & \textbf{100.56}$_{{\color{gain}+0.56}}$\\
\hline
\hline
\multicolumn{5}{c}{\rule{0pt}{2.3ex} MHFlight dataset} \\
\hline
\parbox[c][6ex][c]{1.8cm}{\raggedright Memory Usage ($\%$)} & 100 & \textbf{7.14} & 27.18 & 42.66 \\
\hline
\parbox[c][6ex][c]{1.8cm}{\raggedright Performance Ratio ($\%$)} & 100 & {91.99}$_{{\color{loss}-8.01}}$ & \textbf{99.78}$_{{\color{loss}-0.22}}$ & 99.46$_{{\color{loss}-0.54}}$ \\
\noalign{\hrule height 1pt}
\end{tabular}
\end{table}

Our method leverages only a small subset of the database for retrieval, substantially reducing memory consumption. To quantify this, we report the memory footprint and retrieval performance of various methods in Tab.~\ref{tab:HE_VPR_evaluation_2}. The baseline method, denoted as \textit{Full}, uses the VPR adapter to process the entire database and is assigned 100\% memory usage.
Since the number of sub-databases selected by the HE adapter branch varies per query, all reported values are averaged over the dataset. We also present the relative performance of each method compared to the baseline, where \textit{Full Performance} is defined as the sum of R@1, R@5, and R@10 metrics from full-database retrieval. The \textit{Performance Ratio} is computed as the ratio between the performance of our method and the baseline, with gains and losses indicated by green and red subscripts, respectively.
Results on the GEStudio dataset show that as height estimation improves, the required database size decreases. Using only top-1 candidates, our method achieves approximately 90\% of the baseline performance with just 10\% of the memory. For optimal performance, selecting top-5 or top-10 candidates further boosts accuracy while still requiring significantly less memory --- less than half of what the baseline consumes. These findings highlight the efficiency advantage of HE-VPR in both retrieval quality and resource usage.

\section{Conclusion}
\label{sec:conclusion}

In this paper, we present HE-VPR, an efficient system designed for UAVs operating under dynamic and varying heights. 
By deploying two independent parallel bypass adaptation branches, our framework effectively decouples height estimation and place recognition into a two-stage pipeline while sharing a single frozen foundation backbone. 
This modular architecture successfully mitigates severe scale variations through retrieval-based sub-database selection and center-weighted feature masking, significantly reducing memory consumption and search space without sacrificing retrieval accuracy. 
While the current reliance on discrete height partitions presents a limitation when dealing with continuous height variations, the proposed parameter-efficient design ensures flexibility and provides a robust foundation for future continuous height-adaptive VPR research on aerial platforms.

\typeout{} 
{
\bibliographystyle{IEEEtran}
\bibliography{references,ieeecontrol}
}

\end{document}